\documentclass[10pt,letterpaper]{article}
\title{Confounding variables may degrade generalization performance of radiological deep learning models: a cross-sectional study }
\usepackage[top=0.85in,left=0.85in,footskip=0.85in]{geometry}
\usepackage{hyperref}    

\usepackage{amsmath,amssymb}

\usepackage{changepage}

\usepackage[utf8x]{inputenc}

\usepackage{textcomp,marvosym}

\usepackage{cite}

\usepackage{nameref}

\usepackage{microtype}
\DisableLigatures[f]{encoding = *, family = * }

\usepackage[table]{xcolor}

\usepackage{array}

\newcolumntype{+}{!{\vrule width 2pt}}

\usepackage{multirow}
\usepackage[normalem]{ulem}
\useunder{\uline}{\ul}{}

\usepackage{graphicx}
\usepackage{subcaption}

\usepackage{float}
\usepackage{wrapfig}

\usepackage[colorinlistoftodos]{todonotes}

\newlength\savedwidth

\raggedright
\usepackage[aboveskip=1pt,labelfont=bf,labelsep=period,justification=raggedright,singlelinecheck=off]{caption}

\usepackage{lastpage,fancyhdr,graphicx}
\usepackage{epstopdf}

\fancyheadoffset[L]{2.25in}
\fancyfootoffset[L]{2.25in}
\usepackage{longtable}

\begin{document}
\vspace*{0.2in}

\begin{flushleft}
{\Large
\textbf\newline{Confounding variables can degrade generalization performance of radiological deep learning models} 
}
\newline
\\
John R. Zech\textsuperscript{1*},
Marcus A. Badgeley\textsuperscript{2*},
Manway Liu\textsuperscript{2},
Anthony B. Costa\textsuperscript{3},
Joseph J. Titano\textsuperscript{4},
Eric K. Oermann\textsuperscript{3}
\\
\bigskip

\textbf{1} Department of Medicine, California Pacific Medical Center, San Francisco, CA 94115
\\\texttt{jrz2111@columbia.edu}
\\
\textbf{2} Verily Life Sciences, 269 E Grand Ave, South San Francisco, CA 94080
\\\texttt{marcus.badgeley@icahn.mssm.edu, manwayl@verily.com}
\\
\textbf{3} Department of Neurological Surgery, Icahn School of Medicine, New York, NY 10029
\\\texttt{anthony.costa@mountsinai.org, eric.oermann@mountsinai.org}
\\
\textbf{4} Department of Radiology, Icahn School of Medicine, New York, NY 10029
\\\texttt{joseph.titano@mountsinai.org}
\bigskip

%
%

* These authors contributed equally to this work \\

\end{flushleft}

\section*{Author summary}

Early results in using convolutional neural networks (CNNs) on x-rays to diagnose disease have been promising, but it has not yet been shown that models trained on x-rays from one hospital or one group of hospitals will work equally well at different hospitals. Before these tools are used for computer-aided diagnosis in real-world clinical settings, we must verify their ability to generalize across a variety of hospital systems.

\bigskip

\noindent A cross-sectional design was used to train and evaluate pneumonia screening CNNs on 158,323 chest x-rays from NIH (n=112,120 from 30,805 patients), Mount Sinai (42,396 from 12,904 patients), and Indiana (n=3,807 from 3,683 patients). In 3 / 5 natural comparisons, performance on chest x-rays from outside hospitals was significantly lower than on held-out x-rays from the original hospital systems. CNNs were able to detect where an x-ray was acquired (hospital system, hospital department) with extremely high accuracy and calibrate predictions accordingly.

\bigskip

\noindent The performance of CNNs in diagnosing diseases on x-rays may reflect not only their ability to identify disease-specific imaging findings on x-rays, but also their ability to exploit confounding information. Estimates of CNN performance based on test data from hospital systems used for model training may overstate their likely real-world performance. 

\section*{Abstract}

\textbf{Background}: There is interest in using convolutional neural networks (CNNs) to analyze medical imaging to provide computer aided diagnosis (CAD). Recent work has suggested that image classification CNNs may not generalize to new data as well as previously believed. We assessed how well CNNs generalized across three hospital systems for a simulated pneumonia screening task.

\bigskip

\noindent \textbf{Methods and Findings}: A cross-sectional design with multiple model training cohorts was used to evaluate model generalizability to external sites. 158,323 chest radiographs were drawn from three institutions: NIH (112,120 from 30,805 patients), Mount Sinai Hospital (MSH; 42,396 from 12,904 patients), and Indiana (IU; 3,807 radiographs from 3,683 patients). These patient populations had age mean (S.D.) 46.9 (16.6), 63.2 (16.5), and 49.6 (17), and percent female 43.5\%, 44.8\%, and 57.1\%, respectively. We assessed individual models using area under the receiver operating characteristic curve (AUC) for radiographic findings consistent with pneumonia and compared performance on different test sets with DeLong’s test. The prevalence of pneumonia was high enough at MSH (34.2\%) relative to NIH and IU (1.2\% and 1.0\%) that merely sorting by hospital system achieved an AUC of 0.861 on the joint MSH-NIH dataset. Models trained on data from either NIH or MSH had equivalent performance on IU (p-values 0.580 and 0.273, respectively) and inferior performance on data from each other relative to an internal test set (i.e., new data from within the hospital system used for training data; p-values both \textless\ 0.001). The highest internal performance was achieved by combining training and test data from MSH and NIH (AUC 0.931, 95\% C.I. 0.927-0.936), but this model demonstrated significantly lower external performance at IU (AUC 0.815, 95\% C.I. 0.745-0.885, P = 0.001). To test the effect of pooling data from sites with disparate pneumonia prevalence, we used stratified subsampling to generate MSH-NIH cohorts that only differed in disease prevalence between training data sites. When both training data sites had the same pneumonia prevalence, the model performed consistently on external IU data (P = 0.88). When a ten-fold difference in pneumonia rate was introduced between sites, internal test performance improved compared to the balanced model (10x MSH risk P \textless\ 0.001; 10x NIH P = 0.002), but this outperformance failed to generalize to IU (MSH 10x P \textless\ 0.001; NIH 10x P = 0.027). CNNs were able to directly detect hospital system of a radiograph for 99.95\% NIH (22,050/22,062) and 99.98\% MSH (8,386/8,388) radiographs. The primary limitation of our approach and the available public data is that we cannot fully assess what other factors might be contributing to hospital system-specific biases.

\bigskip

\noindent \textbf{Conclusions}: Pneumonia screening CNNs achieved better internal than external performance in 3 / 5 natural comparisons. When models were trained on pooled data from sites with different pneumonia prevalence, they performed better on new pooled data from these sites but not on external data. CNNs robustly identified hospital system and department within a hospital which can have large differences in disease burden and may confound disease predictions.

\section*{Introduction}

\noindent There is significant interest in using convolutional neural networks (CNNs) to analyze radiology, pathology, or clinical imaging for the purposes of computer aided diagnosis (CAD) \cite{Wang2017-yh,Rajpurkar2017-nh,Gulshan2016-ba,Ting2017-na,Kermany2018-mh}. These studies are generally performed utilizing CNN techniques that were pioneered on well characterized computer vision datasets including the ImageNet Large Scale Visual Recognition Competition (ILSVRC) and the Modified National Institute of Standards and Technology (MNIST) database of hand drawn digits \cite{Russakovsky2014-qx,LeCun1998}. Training CNNs to classify images from these datasets is typically done by splitting the dataset into three subsets: train (data directly used to learn parameters for models), tune (data used to choose hyperparameter settings, also commonly referred to as ‘validation’), and test (data used exclusively for performance evaluation of models learned using train and tune data). CNNs are trained to completion with the first two, and the final set is used to estimate the model’s expected performance on new, previously unseen data. 

\bigskip

\noindent An underlying premise of the test set implying future generalizability to new data is that the test set is reflective of the data that will be encountered elsewhere.  Recent work in computer vision has demonstrated that the true generalization performance of even classic CIFAR-10 photograph classification CNNs to new data may be lower than previously believed \cite{Recht2018-xc}. In the biomedical imaging context, we can contrast ‘internal’ model performance on new, previously unseen data gathered from the same hospital system(s) used for model training with ‘external’ model performance on new, previously unseen data from different hospital systems \cite{Rothwell2005-yp,Pandis2017-zl}. External test data may be different in important ways from internal test data, and this may affect model performance, particularly if confounding variables exist in internal data that do not exist in external data \cite{Cabitza2017-nh}. In a large scale deep learning study of retinal fundoscopy, Ting et al. (2017) noted variation in performance of CNNs trained to identify ocular disease across external hospital systems, with area under the receiver operating characteristic curve (AUC) ranging from 0.889 to 0.983 and image-level concordance with human experts ranging from 65.8\% to 91.2\% on external datasets \cite{Ting2017-na}. Despite the rapid push to develop deep learning systems on radiological data for academic and commercial purposes, to date, no study has looked at whether radiological CNNs actually generalize to external data. If external test performance of a system is inferior to internal test performance, clinicians may erroneously believe systems to be more accurate than they truly are in the deployed context, creating the potential for patient harm.

\bigskip

\noindent The primary aim of this study was to obtain data from three separate hospital systems and to assess how well deep learning models trained at one hospital system generalized to other external hospital systems. For the purposes of this assessment, we chose the diagnosis of pneumonia on chest x-ray for both its clinical significance as well as common occurrence and significant interest \cite{Rajpurkar2017-nh}. By training and testing models on different partitions of data across three distinct institutions, we sought to establish whether a truly generalizable model could be learned as well as which factors affecting external validity could be identified to aid clinicians when assessing models for potential clinical deployment.

\section*{Methods}
\subsection*{Datasets}
This study was approved by the Mount Sinai Health System Institutional Review Board; the requirement for patient consent was waived for this retrospective study that was deemed to carry minimal risk. Three datasets were obtained from different hospital groups: National Institutes of Health Clinical Center (NIH; 112,120 radiographs from 1992-2015), Indiana University Network for Patient Care (IU; 7,470 radiographs, date range not available), and Mount Sinai Hospital (MSH; 48,915 radiographs from 2009-2016) \cite{Wang2017-yh,Demner-Fushman2016-rv}. This study did not have a prospective analysis plan, and all analyses performed are subsequently described.

\subsection*{Convolutional Neural Networks (CNNs)}
Deep learning encompasses any algorithm that uses multiple layers of feed-forward neural networks to model phenomena \cite{LeCun2015-tc}. Classification CNNs are a type of supervised deep learning model that take an image as input and predict the probability of predicted class membership as output. A typical use of CNNs is classifying photographs according to the animals or objects they contain: a chihuahua, a stove, a speedboat, etc. \cite{Russakovsky2014-qx}. Many different CNN architectures have been proposed, including ResNet-50 and DenseNet-121 used in this paper, and improving the performance of these models is an active area of research \cite{He2015-ve,Huang2016-ct}. In practice, CNNs are frequently pre-trained on large computer vision databases, such as ImageNet, rather than being randomly initialized and trained de novo. After pre-training, the CNNs are then fine-tuned on the dataset of interest. This process of pre-training followed by fine-tuning reduces training time, promotes model convergence, and can regularize the model to reduce overfitting. A difficulty of using these models is that there are few formal guarantees as to their generalization performance \cite{Zhang2016-ex}. In this paper, we use CNNs both to preprocess and to predict pneumonia in radiographs.

\subsection*{Preprocessing: Frontal View Filtering}
\begin{figure}[H]
\centering
\caption{Preprocessing approach.}
\includegraphics[width=0.66\textwidth]{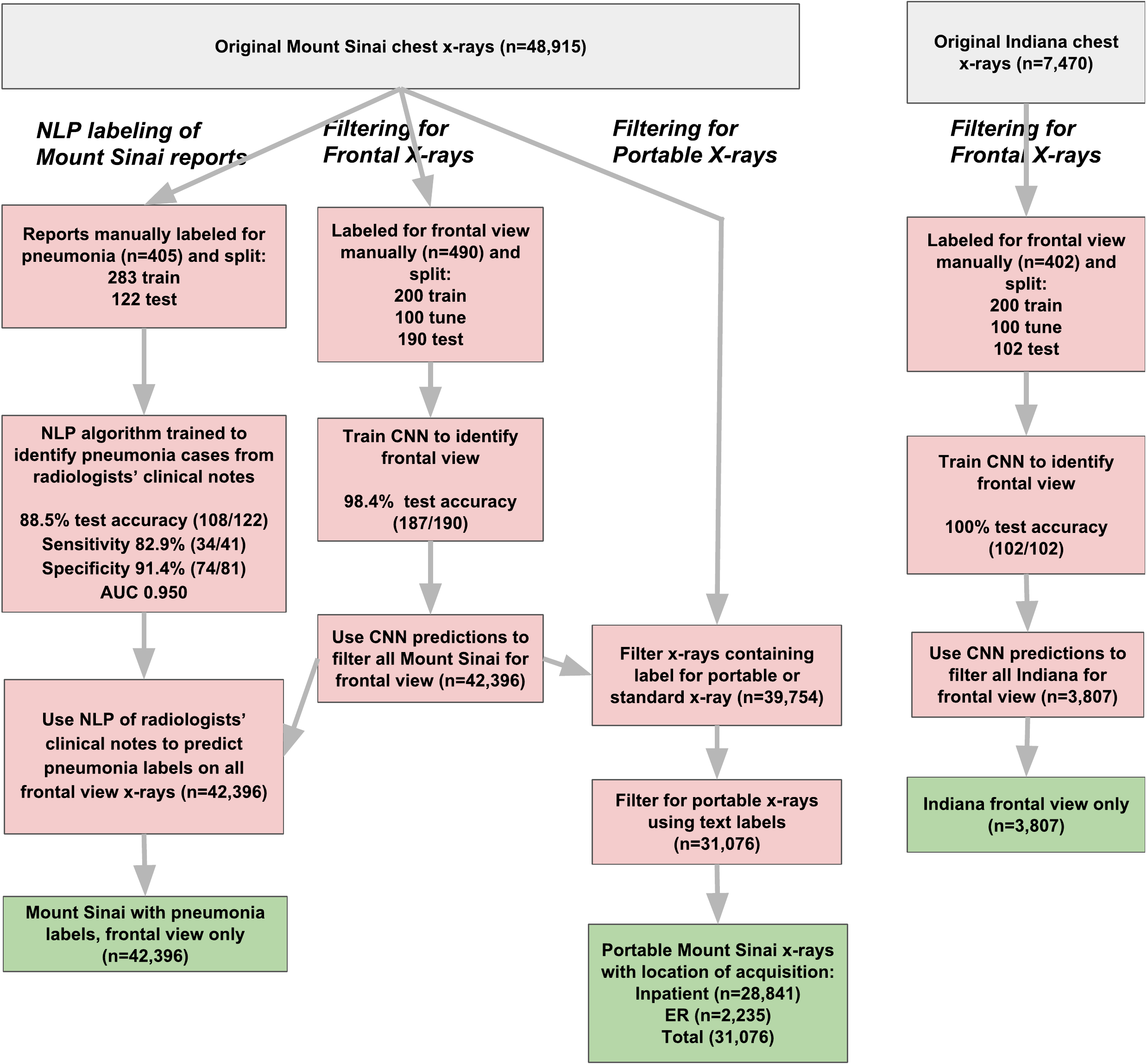}
\label{fig:FigS1}
\end{figure}

NIH data contained only frontal chest radiographs, while IU and MSH data contained both frontal and lateral chest radiographs and were found to contain inconsistent frontal and lateral labels on manual review. 402 IU and 490 MSH radiographs were manually labeled as frontal/lateral and randomly divided into groups (IU: 200 train, 100 tune, 102 test; MSH: 200 train, 100 tune, 190 test) and used to train ResNet-50 CNNs to identify frontal radiographs [14]. 187/190 MSH and 102/102 IU test radiographs were accurately classified. The datasets were then filtered to frontal radiographs using these CNNs, leaving a total of 158,323 radiographs (112,120 NIH, 42,396 MSH, and 3,807 IU) available for analysis (Figure ~\ref{fig:FigS1}).

\subsection*{Preprocessing: Generating Labels for Pathology}
IU radiographs were manually labeled by curators after review of the accompanying text radiology reports \cite{Demner-Fushman2016-rv}. NIH radiographs were labeled automatically using a proprietary natural language processing (NLP) system based on expanding sentences as parse trees and using hand-crafted rules based on the MESH vocabulary to identify statements indicating positive pathology \cite{Rajpurkar2017-nh}. 

\bigskip

\noindent MSH radiographs did not initially include labels, so a subset of radiographic reports were manually labelled to train an NLP algorithm that could infer labels for the full dataset. 405 radiographic reports were manually labeled for cardiomegaly, emphysema, effusion, hernia, nodule, atelectasis, pneumonia, edema, and consolidation. To evaluate the NLP algorithm’s performance, these were split into train and test groups (283 and 122, respectively). A previously described NLP concept extraction model based on 1- and 2-gram bag-of-words features with Lasso logistic regression was trained to identify reports positive for these findings \cite{Zech2018-tg}.  AUC, sensitivity, and specificity at a 50\% classification threshold are reported in in Table ~\ref{TableS1}. The NLP model was then refit with all 405 manually labelled reports and used to process all unlabelled reports. As reports positive for hernia occurred too infrequently to use this NLP algorithm, reports were automatically labeled as positive for hernia if the word ‘hernia’ appeared in the report.

\begin{table}[H]
\centering
\caption{Performance of NLP algorithm on 30\% test data.}
\label{TableS1}
\begin{tabular}{|l|c|c|c|c}
\cline{1-4}
\textbf{Finding}       & \textbf{AUC} & \textbf{Sensitivity} & \textbf{Specificity} &  \\ \cline{1-4}
\textbf{Cardiomegaly}  & 0.97         & 1.00 (34/34)         & 0.92 (81/88)         &  \\ \cline{1-4}
\textbf{Emphysema}     & 0.85         & 0.833 (5/6)          & 1.00 (116/116)       &  \\ \cline{1-4}
\textbf{Effusion}      & 0.98         & 0.918 (45/49)        & 0.945 (69/73)        &  \\ \cline{1-4}
\textbf{Atelectasis}   & 0.99         & 0.971 (34/35)        & 0.989 (86/87)        &  \\ \cline{1-4}
\textbf{Pneumonia}     & 0.95         & 0.829 (34/41)        & 0.914 (74/81)        &  \\ \cline{1-4}
\textbf{Edema}         & 0.95         & 0.875 (21/24)        & 0.939 (92/98)        &  \\ \cline{1-4}
\textbf{Consolidation} & 0.97         & 0.891 (57/64)        & 0.897 (52/58)        &  \\ \cline{1-4}
\textbf{Nodule}        & 1.00         & 1.00 (3/3)           & 1.00 (119/119)       &  \\ \cline{1-4}
\end{tabular}
\end{table}

\subsection*{Preprocessing: Separation of Patients Across Train / Tune / Test Groups}
\begin{figure}[H]
\caption{Cohort splitting diagram.}
\includegraphics[width=\textwidth]{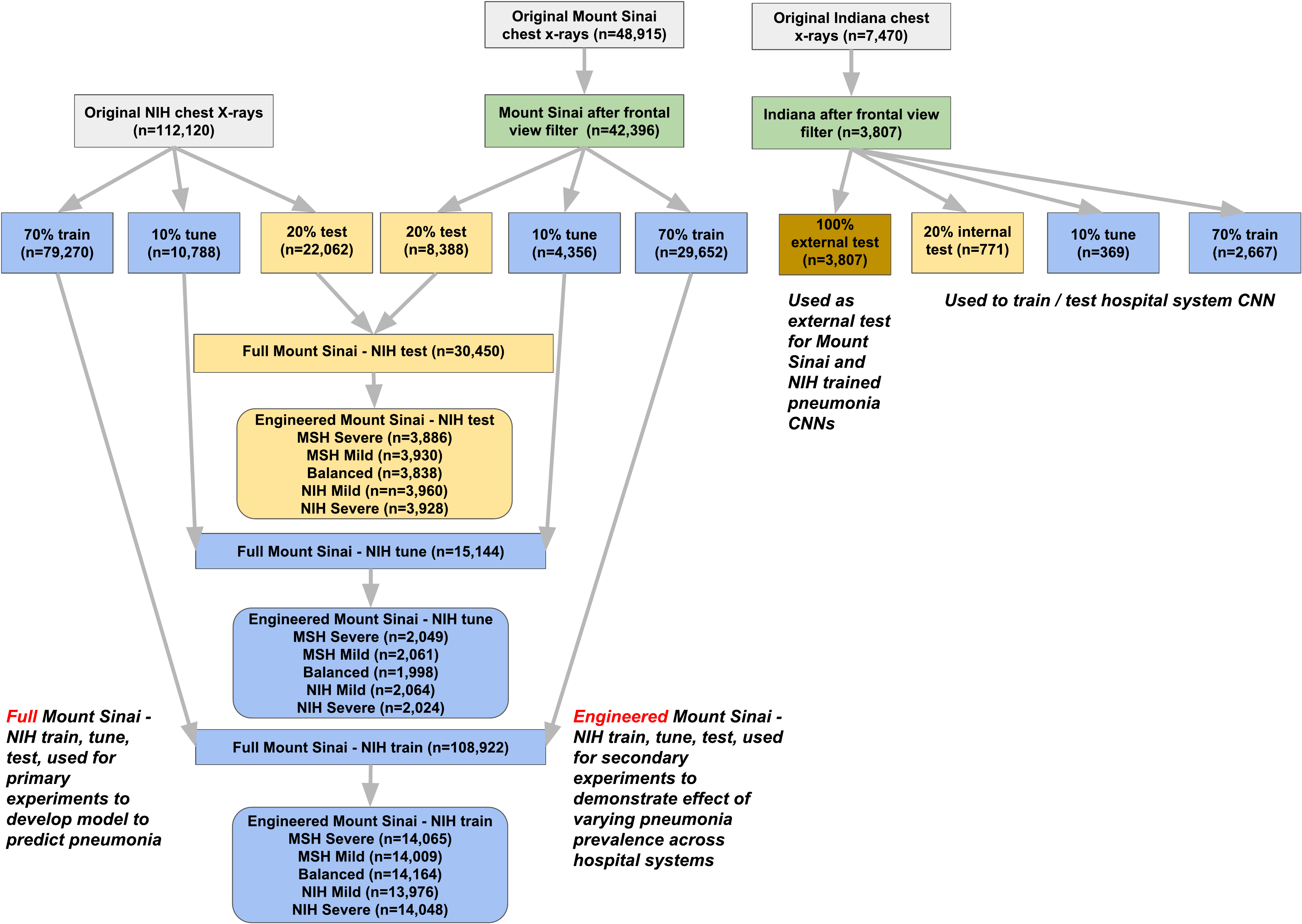}
\label{fig:FigS2}
\end{figure}

As NIH and MSH data contained patient identifiers, all NIH patients were separated into fixed train (70\%), tune (10\%), and test (20\%) groups (Figure ~\ref{fig:FigS2}). IU data did not contain patient identifiers. In the case of pneumonia detection, 100\% of IU data was reserved for use as an external test set. IU data was used for training only to detect hospital system, and in this case was separated into fixed train (70\%), tune (10\%), and test (20\%) groups using an identifier corresponding to accession number (e.g., which radiographs were obtained at the same time on the same patient). Test data was not available to CNNs during model training, and all results reported in this study are calculated exclusively on test data. 

\subsection*{Preprocessing: Identifying Mount Sinai Portable Scans From Inpatient Wards and Emergency Department}
Of 42,396 MSH radiographs, 39,574 contained a label indicating whether they were portable radiographs; 31,838 were labeled as portable. We identified a subset of 31,076 MSH portable radiographs that documented the department of acquisition, with 28,841 from inpatient wards and 2,235 from the emergency department.

\subsection*{Model Training}
PyTorch 0.2.0 and torchvision were used for model training \cite{Paszke2017-jo}. All images were resized to 224 x 224. CNNs used for experiments were trained with DenseNet-121 architecture with an additional dense layer (n=15) attached to the original bottleneck layer and sigmoid activation (for binary classification) or a linear layer with output dimension equal to that of the classification label followed by softmax activation (for n\textgreater2 multiclass prediction) \cite{Huang2016-ct}. This additional dense layer was added to facilitate extraction of bottleneck features in a reduced dimension. A DenseNet architecture with weights pretrained to ImageNet was chosen to facilitate comparison with recent work on pneumonia detection in radiographs by Rajpurkar et al. and for its state-of-the-art results on standard computer vision datasets \cite{Rajpurkar2017-nh}. All models were trained using a cross entropy loss function with parameter update by stochastic gradient descent with momentum, initial learning rate 0.01, momentum 0.9, weight decay 0.0001. Learning rate was decayed by a factor of 10 after each epoch with no improvement in validation loss, and training was stopped after 3 epochs with no improvement in validation loss. 

\subsection*{Internal and External Performance Testing}
To assess how individual models trained using single datasets would generalize compared to a model trained simultaneously on multiple datasets, we trained CNNs to predict nine overlapping diagnoses (cardiomegaly, emphysema, effusion, hernia, nodule, atelectasis, pneumonia, edema, and consolidation) using 3 different train set combinations: NIH, MSH, and a joint NIH-MSH train set. We were interested only in the prediction of pneumonia and included other diagnoses to improve overall model training and performance. For each model, we calculated AUC, accuracy, sensitivity, and specificity for 4 different test sets: joint NIH-MSH, NIH only, MSH only, and IU. We report differences in test AUC for all possible internal-external comparisons. We consider the joint MSH-NIH test set the internal comparison set for the jointly trained model. We additionally report differences in test AUC between a jointly trained MSH-NIH model and individual MSH-NIH test sets. Classification threshold was set to ensure 95\% sensitivity on each test set to simulate model use for a theoretical screening task. After external review of this analysis, a trivial model that ranked cases based only on the average pneumonia prevalence in each hospital system’s training data and completely ignored radiographic findings was evaluated on the MSH-NIH test set to evaluate how hospital system alone can predict pneumonia in the joint dataset.

\subsection*{Hospital System and Department Prediction}
After training models for pneumonia and evaluating their performance across sites, additional analysis was planned to better understand a CNN’s ability to detect site and department and how that could affect pneumonia prediction. We trained a CNN to predict hospital system from radiographs to assess whether location information was directly detectable from the radiograph alone. Radiographs from all three hospital systems were utilized to learn a model that could identify the hospital system from which a given radiograph was drawn. To develop this concept more granularly, for MSH radiographs, we further identified from which hospital unit individual radiographs were obtained (inpatient wards, emergency department). In all cases, we report the classification accuracy on a held-out test set.   

\subsection*{Sample Activation Maps}
We created 7x7 sample activation maps following Zhou et al. (2015) to attempt to understand which locations in chest radiographs provided strong evidence for hospital system \cite{Zhou2015-ho}. For this experiment, we specifically identify radiographs from the NIH. For a sample of NIH test radiographs (n=100) we averaged the softmax probability for each subregion calculated as \bigskip

\( P \left( hospital==NIH \vert  radiograph_{i,j} \right) =\frac{e^{Y_{i,j NIH}}}{e^{Y_{i,j NIH}}+e^{Y_{i,j MSH}}+e^{Y_{i,j IU}}} \)\bigskip
 
\noindent where $i,j$ corresponds to the subregion at the $i$th row and $j$th column of the final convolutional layer (7x7 = 49 subregions), where each 
\bigskip

 \( Y_{i,j \:Hospital\:System}= \sum _{1}^{K} \left( \beta_{k\:Hospital \:System}*X_{k,i,j} \right) +\beta_{0 \:Hospital \:System} \)
 
\bigskip
\noindent where the sum is performed over the $K$ final convolutional layers and $X_{k,i,j}$ represents the activation at the $i$th row and $j$th column of the $k$th final convolutional layer. To characterize how many different subregions were typically involved in NIH hospital system classification, we report the mean, minimum, and maximum number of subregions that predicted NIH decisively (probability \textgreater= 95\%). To illustrate the contribution of particularly influential features (e.g., laterality labels) to classification, we present several examples of heatmaps generated by calculating $Y_{i,j NIH}-Y_{i,j MSH}-Y_{i,j IU}$ for all $i,j$ subregions in an image and subtracting the mean. This additional calculation was necessary to distinguish their positive contribution in the context of many subregions contributing positively to classification probability.

\subsection*{Engineered Relative Risk Experiment}
We wished to assess the hypothesis that a CNN could learn to exploit large differences in pathology prevalence between two hospital systems in training data by calibrating its predictions to the baseline prevalence at each hospital system, rather than exclusively discriminating based on direct pathology findings. This would lead to strong performance on a test dataset consisting of imbalanced data from both hospital systems but would fail to generalize to data from an external hospital system. To test this hypothesis, we simulated experimental cohorts that differed only in relative disease prevalence and performed internal and external evaluations as described above. Five cohorts of 20,000 patients consisting of 10,000 MSH and 10,000 NIH patients were specifically sampled to artificially set different levels of pneumonia prevalence in each population, while maintaining a constant overall prevalence: NIH Severe (NIH 9.9\%, MSH 0.1\%), NIH Mild (NIH 9\%, MSH 1\%), Balanced (NIH 5\%, MSH 5\%), MSH Mild (MSH 9\%, NIH 1\%), MSH Severe (MSH 9.9\%, NIH 0.1\%). The sampling routine also ensured that males and females had equal prevalence of pneumonia. We refer to these as ‘engineered prevalence cohorts.’ Train / tune / test splits consistent with prior modeling were maintained for these experiments. CNNs were trained on each cohort in the fashion previously described, and test AUCs on internal joint MSH-NIH and external IU data were compared.

\subsection*{Statistical Methods}
To assess differences between classification models, we used either the paired or unpaired version of DeLong’s test for ROC curves as appropriate \cite{DeLong1988-je}. Comparisons between proportions were performed utilizing $\chi^2$ tests, and all p-values were assessed at an alpha of 0.05. Statistical analysis was performed using R version 3.4 with the pROC package and scikit-learn 0.18.1 \cite{Robin_undated-jg,Pedregosa2011-qq}.



\section*{Results}

\subsection*{Datasets}

The average age of patients in the MSH cohort was 63.2 years (S.D. 16.5), compared to 49.6 years (S.D. 17 years) in the IU cohort, and 46.9 years (S.D. 16.6 years) in the NIH cohort (Table ~\ref{Table1}). Positive cases of pneumonia were remarkably more prevalent in MSH data (34.2\%) than in either NIH (1.2\%, $\chi^2$ P \textless\ 0.001) or IU (1.0\%, P \textless\ 0.001) data. 

\begin{table}[H]
\centering
\caption{Baseline characteristics of datasets by hospital system.}
\label{Table1}
\begin{tabular}{|l|c|c|c|}
\hline
\textbf{Characteristic}                               & \textbf{IU}                           & \textbf{MSH}                           & \textbf{NIH}                             \\ \hline
Patient demographics                                  &                                            &                                                &                                          \\ \hline
No. patient radiographs                               & 3,807                                      & 42,396                                         & 112,120                                  \\ \hline
No. patients                                          & 3,683                                      & 12,904                                         & 30,805                                   \\ \hline
Age, mean (SD), years                                 & 49.6 (17.0)                                & 63.2 (16.5)                                    & 46.9 (16.6)                              \\ \hline
No. females (\%)                                      & 643 (57.1)                                 & 18,993 (44.8\%)                                & 48,780 (43.5\%)                          \\ \hline
Image diagnosis frequencies                           &                                            &                                                &                                          \\ \hline
Pneumonia, No. (\%)                                   & 39 (1.0\%)                                 & 14,515 (34.2\%)                                & 1,353 (1.2\%)                            \\ \hline
Emphysema, No. (\%)                                   & 62 (1.6\%)                                 & 1,308 (3.1\%)                                  & 2,516 (2.2\%)                            \\ \hline
Effusion, No. (\%)                                    & 142 (3.7\%)                                & 19,536 (46.1\%)                                & 13,307 (11.9\%)                          \\ \hline
Consolidation, No. (\%)                               & 26 (0.7\%)                                 & 25,318 (59.7\%)                                & 4,667 (4.2\%)                            \\ \hline
Nodule, No. (\%)                                      & 104 (2.7\%)                                & 569 (1.3\%)                                    & 6,323 (5.6\%)                            \\ \hline
Atelectasis, No. (\%)                                 & 307 (8.1\%)                                & 16,713 (39.4\%)                                & 11,535 (10.3\%)                          \\ \hline
Edema, No. (\%)                                       & 45 (1.2\%)                                 & 7,144 (16.9\%)                                 & 2,303 (2.1\%)                            \\ \hline
Cardiomegaly, No. (\%)                               & 328 (8.6\%)                                & 14,285 (33.7\%)                                & 2,772 (2.5\%)                            \\ \hline
Hernia, No. (\%)                                      & 46 (1.2\%)                                 & 228 (0.5\%)                                    & 227 (0.2\%)                              \\ \hline
\multicolumn{4}{|l|}{\begin{tabular}[c]{@{}l@{}}*Sex data available for 1,122 / 3,807 Indiana, 42,383 / 42,396 Mount Sinai; age\\   data available for 112,077 / 112,120 NIH\end{tabular}} \\ \hline
\end{tabular}
\end{table}

\subsection*{Internal and External Performance Testing}

\begin{wrapfigure}[25]{r}{0.50\textwidth}
\caption{Pneumonia models evaluated on internal and external test sets. A model trained using both Mount Sinai and NIH data (MSH+NIH) had higher performance on the combined MSH+NIH test set than on either subset individually or on fully external Indiana (IU) data.}
\centering
\includegraphics[trim=80 07 100 35,scale=0.71]{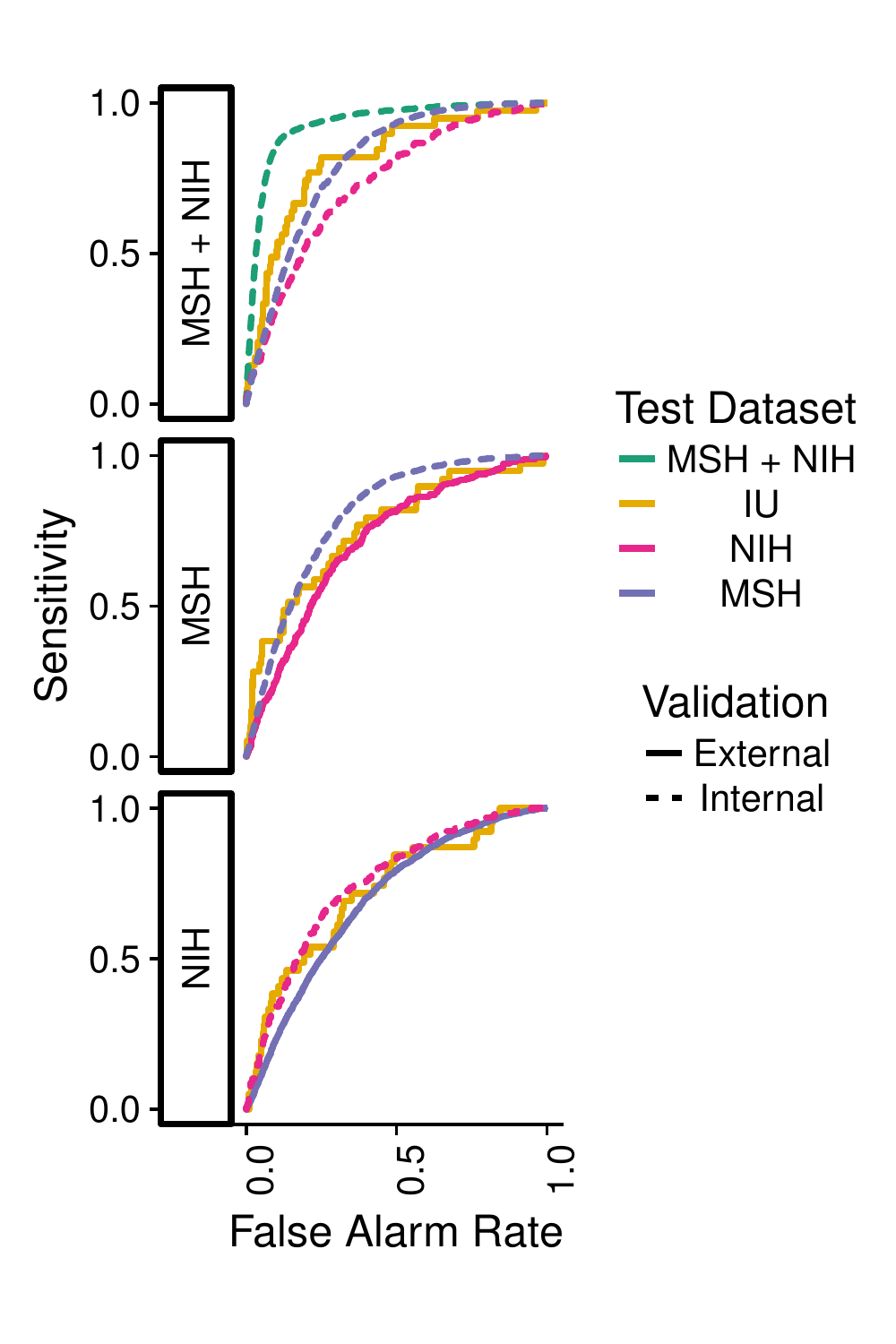}
\label{fig:Fig1}
\end{wrapfigure}

The internal performance of pneumonia detection CNNs significantly exceeded external performance in 3 / 5 natural comparisons (Figure ~\ref{fig:Fig1}, Table ~\ref{Table2}). CNNs trained to detect pneumonia at NIH had internal test AUC 0.750 (95\% C.I. 0.721-0.778), significantly worse external test AUC 0.695 at MSH (95\% C.I. 0.683-0.706, P \textless\ 0.001), and comparable external test AUC 0.725 at IU (95\% C.I. 0.644-0.807, P = 0.580). CNNs trained to detect pneumonia at MSH had internal test AUC 0.802 (95\% C.I. 0.793-0.812), significantly worse external test AUC 0.717 at NIH (95\% C.I. 0.687-0.746, P \textless\ 0.001) and comparable external test AUC 0.756 at IU (95\% C.I. 0.674-0.838, P = 0.273). A jointly trained MSH-NIH model had internal test AUC 0.931 (95\% C.I. 0.927-0.936), significantly greater than external test AUC 0.815 at Indiana (95\% C.I. 0.745-0.885, P = 0.001). The jointly trained model had stronger internal performance compared to either constituent site individually (MSH AUC 0.805, 95\% C.I 0.796-0.814, P \textless\ 0.001; NIH AUC 0.733, 95\% C.I. 0.703-0.762, P \textless\ 0.001) (Table ~\ref{Table2}). A trivial model that ranked cases based only on the average pneumonia prevalence in each hospital system achieved AUC 0.861 on the joint MSH-NIH test set.

\bigskip

\bigskip

\begin{table}[H]
\centering
\caption{Internal and external pneumonia screening performance for all train - tune and test hospital system combinations.}
\label{Table2}
\resizebox{\textwidth}{!}{\begin{tabular}{|c|c|c|c|c|c|c|c|c|}
\hline
\textbf{\begin{tabular}[c]{@{}c@{}}Train - \\ Tune \\ Site\end{tabular}}         & \textbf{\begin{tabular}[c]{@{}c@{}}Comparison \\ Type*\end{tabular}} & \textbf{\begin{tabular}[c]{@{}c@{}}Test Site \\ (Images)\end{tabular}}             & \textbf{AUC (95\% C.I.)} & \textbf{Acc.} & \textbf{Sens.} & \textbf{Spec.} & \textbf{PPV} & \textbf{NPV} \\ \hline
\multirow{5}{*}{NIH}                                                             & Internal                                                             & NIH (N=22,062)                                                                     & 0.750 (0.721-0.778)      & 0.255         & 0.951          & 0.247          & 0.015        & 0.998        \\ \cline{2-9} 
                                                                                 & External                                                             & MSH (N=8,388)                                                              & 0.695 (0.683-0.706)      & 0.476         & 0.950          & 0.212          & 0.401        & 0.884        \\ \cline{2-9} 
                                                                                 & External                                                             & IU (N=3,807)                                                                  & 0.725 (0.644-0.807)      & 0.190         & 0.974          & 0.182          & 0.012        & 0.999        \\ \cline{2-9} 
                                                                                 & Superset                                                             & MSH + NIH (N=30,450)                                                       & 0.773 (0.766-0.780)      & 0.462         & 0.950          & 0.403          & 0.160        & 0.985        \\ \cline{2-9} 
                                                                                 & Superset                                                             & \begin{tabular}[c]{@{}c@{}}MSH + NIH + IU\\   (N=34,257)\end{tabular} & 0.787 (0.780-0.793)      & 0.470         & 0.950          & 0.418          & 0.148        & 0.987        \\ \hline
\multirow{5}{*}{\begin{tabular}[c]{@{}c@{}}MSH\end{tabular}}          & Internal                                                             & MSH (N=8,388)                                                              & 0.802 (0.793-0.812)      & 0.617         & 0.950          & 0.432          & 0.482        & 0.940        \\ \cline{2-9} 
                                                                                 & External                                                             & NIH (N=22,062)                                                                     & 0.717 (0.687-0.746)      & 0.184         & 0.951          & 0.175          & 0.014        & 0.997        \\ \cline{2-9} 
                                                                                 & External                                                             & IU (N=3,807)                                                                  & 0.756 (0.674-0.838)      & 0.099         & 0.974          & 0.090          & 0.011        & 0.997        \\ \cline{2-9} 
                                                                                 & Superset                                                             & MSH + NIH (N=30,450)                                                       & 0.862 (0.856-0.868)      & 0.562         & 0.950          & 0.516          & 0.190        & 0.989        \\ \cline{2-9} 
                                                                                 & Superset                                                             & \begin{tabular}[c]{@{}c@{}}MSH + NIH + IU\\   (N=34,257)\end{tabular} & 0.871 (0.865-0.877)      & 0.577         & 0.950          & 0.537          & 0.180        & 0.990        \\ \hline
\multirow{5}{*}{\begin{tabular}[c]{@{}c@{}}MSH + \\ NIH\end{tabular}} & Internal                                                             & MSH + NIH (N=30,450)                                                       & 0.931 (0.927-0.936)      & 0.732         & 0.950          & 0.706          & 0.279        & 0.992        \\ \cline{2-9} 
                                                                                 & Subset                                                               & NIH (N=22,062)                                                                     & 0.733 (0.703-0.762)      & 0.243         & 0.951          & 0.234          & 0.015        & 0.997        \\ \cline{2-9} 
                                                                                 & Subset                                                               & MSH (N=8,388)                                                              & 0.805 (0.796-0.814)      & 0.630         & 0.950          & 0.451          & 0.491        & 0.942        \\ \cline{2-9} 
                                                                                 & External                                                             & IU (N=3,807)                                                                  & 0.815 (0.745-0.885)      & 0.238         & 0.974          & 0.230          & 0.013        & 0.999        \\ \cline{2-9} 
                                                                                 & Superset                                                             & \begin{tabular}[c]{@{}c@{}}MSH + NIH + IU\\   (N=34,257)\end{tabular} & 0.934 (0.929-0.938)      & 0.732         & 0.950          & 0.709          & 0.258        & 0.993        \\ \hline
\multicolumn{9}{|c|}{\begin{tabular}[c]{@{}c@{}}*Superset= a test dataset containing data from the same distribution (hospital system) as the training data as \\ well as external data. Subset = a test dataset containing data from fewer distributions (hospital systems) then \\ the training data.\end{tabular}}                                   \\ \hline
\end{tabular}}
\end{table}

\subsection*{Hospital System and Department Prediction}

\begin{figure}[H]
\centering
\caption{CNN to predict hospital system detected both general and specific image features. (\subref{fig:Fig_2a}) We obtained activation heatmaps from our trained model and averaged over a sample of images to reveal which subregions tended to contribute to a hospital system classification decision. Many different subregions strongly predicted the correct hospital system, with especially strong contributions from image corners. (\subref{fig:Fig_2b}-\subref{fig:Fig_2c}) On individual images, which have been normalized to highlight only the most influential regions and not all those that contributed to a positive classification, we note that the CNN has learned to detect a metal token that radiology technicians place on the patient in the corner of the image field of view at the time they capture the image. When these strong features are correlated with disease prevalence, models can leverage them to indirectly predict disease.}
\begin{subfigure}[t]{.34\textwidth}
\centering
\includegraphics[width=\linewidth]{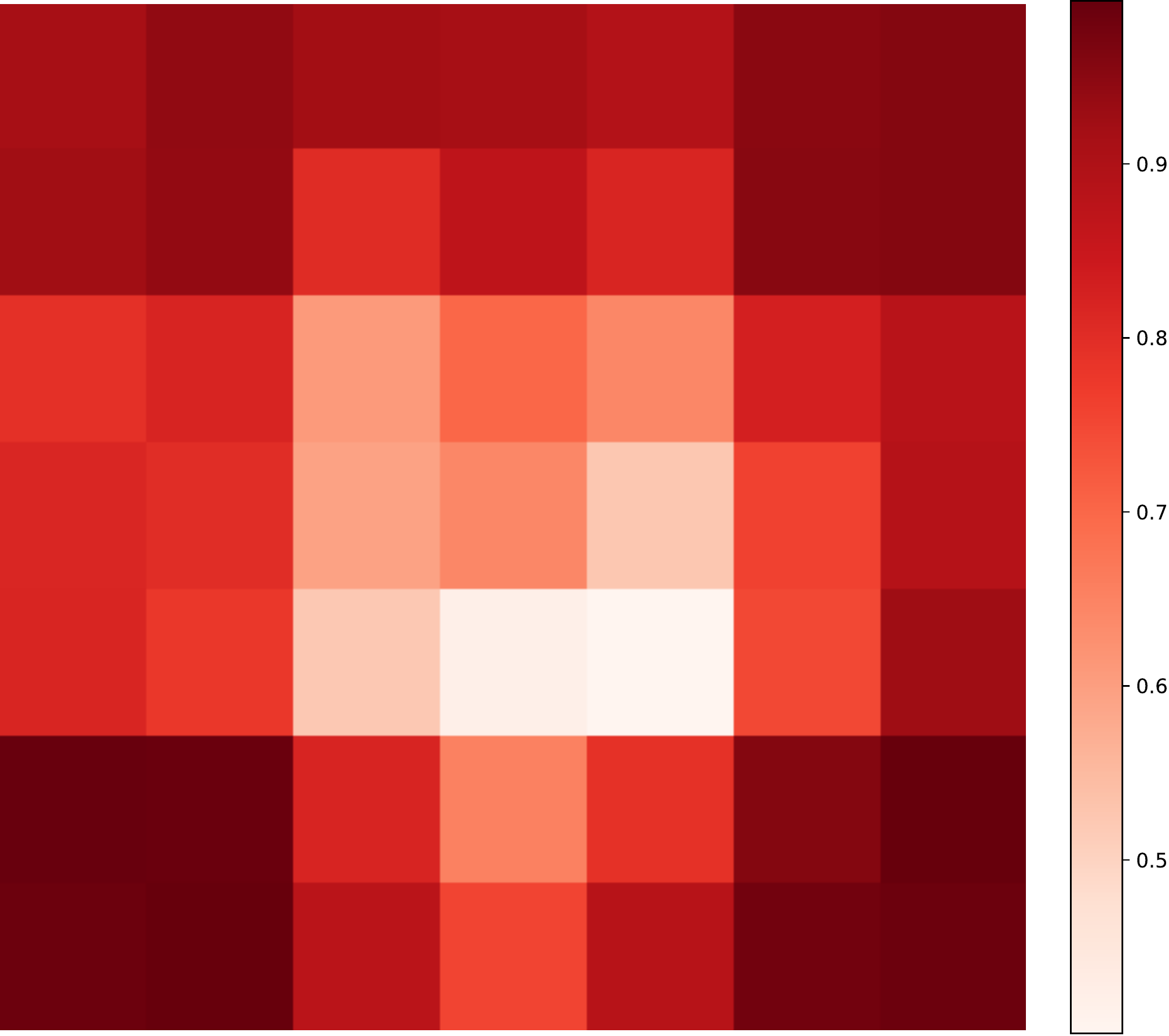}
\caption{}\label{fig:Fig_2a}
\end{subfigure}
\begin{subfigure}[t]{.31\textwidth}
\centering
\includegraphics[width=\linewidth]{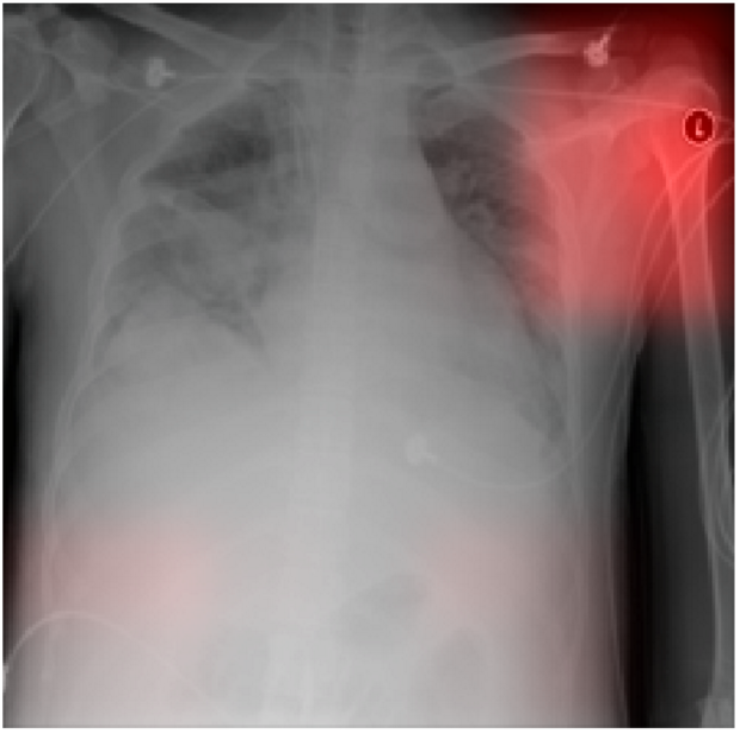}
        \caption{}\label{fig:Fig_2b}
\end{subfigure}
\begin{subfigure}[t]{.31\textwidth}
\centering
\includegraphics[width=\linewidth]{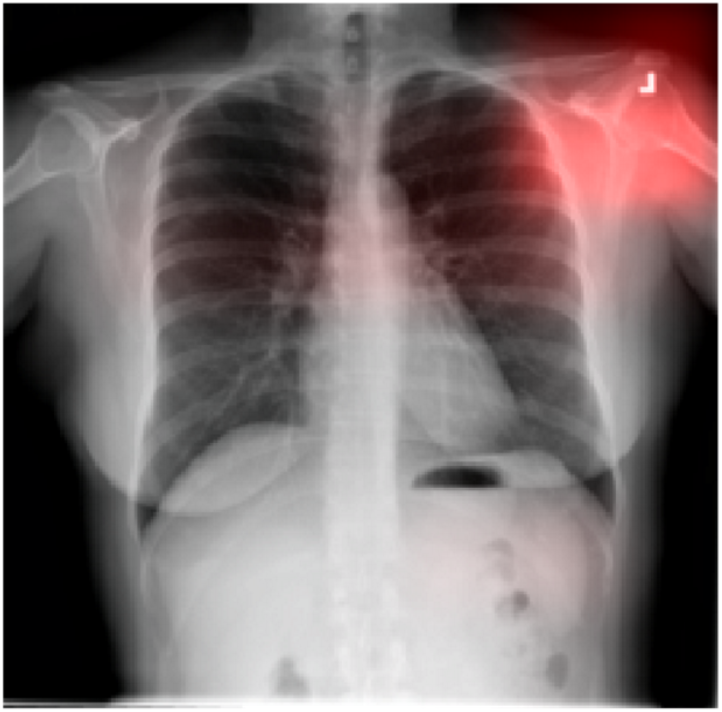}
\caption{}\label{fig:Fig_2c}
\end{subfigure}

\end{figure}

A CNN trained to identify hospital system accurately identified 22,050/22,062 (99.95\%) of NIH, 8,386/8,388 (99.98\%) of MSH, and 737/771 (95.59\%) of IU test radiographs. To identify radiographs originating from a specific hospital system, such as NIH, CNNs used features from many different image regions (Figure ~\ref{fig:Fig_2a}); the majority of image subregions were individually able to predict the hospital system with \textgreater= 95\% certainty (35.7 / 49, 72.9\%, min 21, max 49, N = 100 NIH radiographs). Laterality labels were particularly influential (Figure ~\ref{fig:Fig_2b}-\ref{fig:Fig_2c}).

\bigskip

\noindent A CNN trained to identify individual departments within MSH accurately identified 5,805/5,805 (100\%) of inpatient radiographs and 449/449 (100\%) of emergency department radiographs. Patients who received portable radiographs on an inpatient floor had a higher prevalence of pneumonia than those in the emergency department (41.1\% versus 32.8\%, respectively, $\chi^2$ P \textless\ 0.001).

\bigskip

\subsection*{Engineered Relative Risk Experiment}

\begin{figure}[H]
\caption{Assessing how prevalence differences in aggregated datasets encouraged confounder exploitation. \textbf{(A)} Five cohorts of 20,000 patients engineered to differ only in relative pneumonia risk based on hospital system.  Model performance was assessed on combined test data from the internal hospital systems (MSH+NIH) and separately on test data from an external hospital system (IU). \textbf{(B)} Although models performed better in internal testing in the presence of extreme prevalence differences, this benefit was not seen when applied to data from new hospital systems. The natural relative risk of disease at Mount Sinai (MSH), indicated by a vertical line, was quite imbalanced.}
\centering
\includegraphics[width=0.75\textwidth]{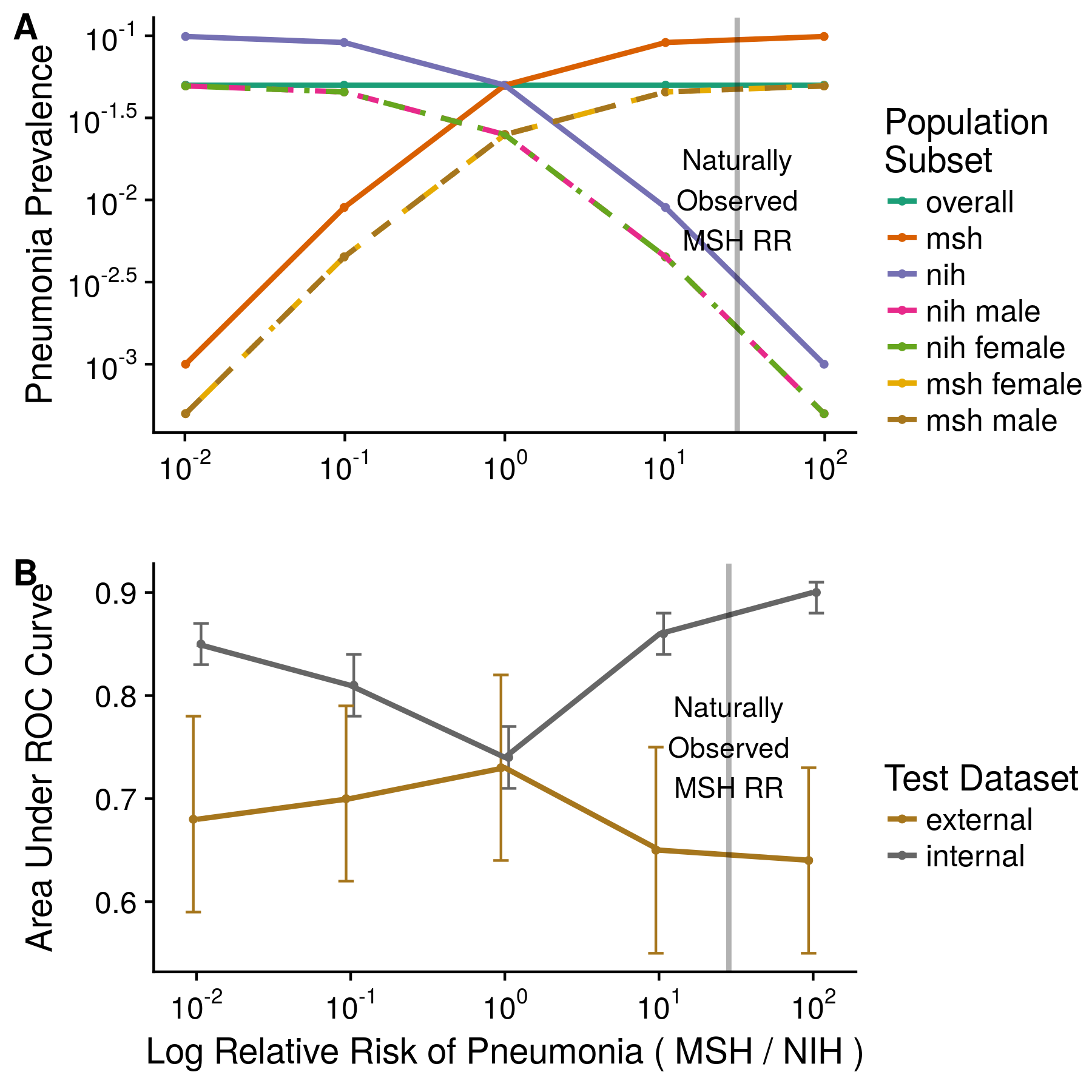}
\label{fig:Fig3}
\end{figure}

Artificially increasing the difference in the prevalence of pneumonia between MSH and NIH led to CNNs that performed increasingly well on internal testing but not external testing (Table ~\ref{TableS2}). CNNs trained on engineered prevalence cohorts of NIH and MSH data showed stronger internal AUC on a joint NIH-MSH test set when the prevalence of pneumonia was imbalanced between the two hospital systems in the training dataset with MSH Severe AUC 0.899 (95\% C.I. 0.885-0.914, P \textless\ 0.001), MSH Mild AUC 0.860 (0.839-0.882, P \textless\ 0.001), NIH Mild AUC 0.807 (95\% C.I. 0.778-0.836, P = 0.002), and NIH Severe AUC 0.849 (95\% C.I. 0.826-0.871, P \textless\ 0.001) than when it was balanced with AUC 0.739 (95\% C.I. 0.707-0.772) (Figure ~\ref{fig:Fig3}). 

\bigskip

\noindent Internal MSH-NIH performance of all models trained on imbalanced cohorts was significantly better than their corresponding external performance on IU (external MSH Severe AUC 0.641, 95\% C.I. 0.552-0.730, P \textless\ 0.001; MSH Mild AUC 0.650, 95\% C.I. 0.548-0.752, P \textless\ 0.001; NIH Mild AUC 0.703, 95\% C.I. 0.616-0.790, P = 0.027; NIH Severe AUC 0.683, 95\% C.I. 0.591-0.775, P \textless\ 0.001). Internal MSH-NIH performance did not significantly exceed external IU performance for Balanced (0.739, 95\% C.I. 0.707-0.772 vs. 0.732, 95\% C.I. 0.645-0.819, P = 0.880).

\begin{table}[H]
\centering
\caption{Internal and external pneumonia screening performance for datasets with engineered pneumonia prevalences.}
\label{TableS2}
\resizebox{0.92\textwidth}{!}{\begin{tabular}{|c|c|c|c|c|c|c|c|}
\hline
\textbf{\begin{tabular}[c]{@{}c@{}}Engineered \\MSH - \\ NIH Cohorts\end{tabular}} & \textbf{Test Site (Images)}   & \textbf{AUC (95\% C.I.)} & \textbf{Acc.} & \textbf{Sens.} & \textbf{Spec.} & \textbf{PPV} & \textbf{NPV} \\ \hline
\multirow{2}{*}{\textbf{MSH Severe}}                      & Internal Engineered (N=3,886) & 0.899 (0.885-0.914)      & 0.690             & 0.953                & 0.674                & 0.146        & 0.996        \\ \cline{2-8} 
                                                                  & External IU (N=3,807)    & 0.641 (0.552-0.730)      & 0.111             & 0.974                & 0.102                & 0.011        & 0.997        \\ \hline
\multirow{2}{*}{\textbf{MSH Mild}}                        & Internal Engineered (N=3,930) & 0.860 (0.839-0.882)      & 0.523             & 0.951                & 0.497                & 0.103        & 0.994        \\ \cline{2-8} 
                                                                  & External IU (N=3,807)    & 0.650 (0.548-0.752)      & 0.050             & 0.974                & 0.041                & 0.010        & 0.994        \\ \hline
\multirow{2}{*}{\textbf{Balanced}}                                & Internal Engineered (N=3,838) & 0.739 (0.707-0.772)      & 0.325             & 0.951                & 0.289                & 0.071        & 0.991        \\ \cline{2-8} 
                                                                  & External IU (N=3,807)    & 0.732 (0.645-0.819)      & 0.057             & 0.974                & 0.048                & 0.010        & 0.994        \\ \hline
\multirow{2}{*}{\textbf{NIH Mild}}                                & Internal Engineered (N=3,960) & 0.807 (0.778-0.836)      & 0.439             & 0.952                & 0.414                & 0.074        & 0.994        \\ \cline{2-8} 
                                                                  & External IU (N=3,807)    & 0.703 (0.616-0.790)      & 0.175             & 0.974                & 0.167                & 0.012        & 0.998        \\ \hline
\multirow{2}{*}{\textbf{NIH Severe}}                              & Internal Engineered (N=3,928) & 0.849 (0.826-0.871)      & 0.572             & 0.954                & 0.552                & 0.100        & 0.996        \\ \cline{2-8} 
                                                                  & External IU (N=3,807)    & 0.683 (0.591-0.775)      & 0.051             & 0.974                & 0.042                & 0.010        & 0.994        \\ \hline
\end{tabular}}
\end{table}

\section*{Discussion}

We have demonstrated that pneumonia screening CNNs trained on data from individual or multiple hospital systems did not consistently generalize to external sites, nor did they make predictions exclusively based on underlying pathology. We note that the issue of not generalizing externally is distinct from typical train/test performance degradation, in which overfitting to training data leads to lower performance on testing data: in our experiments, all results are reported on held-out test data exclusively in both internal and external comparisons. Performance of the jointly trained MSH-NIH model on the joint test set (AUC 0.931) was higher than performance on either individual dataset (AUC 0.805 and 0.733, respectively), likely because the model was able to calibrate to different prevalences across hospital systems in the joint test set by detecting specific features in imaging. For comparison, a simple calibration-based non-CNN model that used hospital system pneumonia prevalence only to make pneumonia predictions and ignored image features achieved AUC 0.861 in the joint MSH-NIH test set due to the large difference in pneumonia prevalence between the MSH and NIH test sets. 

\bigskip

\noindent By engineering cohorts of varying prevalence, we demonstrated that the more pneumonia rates differed between hospital systems, the more they were exploited to make predictions, which led to poor generalization on external datasets. We noted that metallic tokens indicating laterality often appeared in radiographs in a site-specific way which made hospital system identification trivial. However, CNNs did not require this indicator: most image subregions contained features indicative of a radiograph’s origin. These results suggest that CNNs could rely on subtle differences in acquisition protocol, image processing, or distribution pipeline (e.g., image compression) and overlook pathology. This can lead to strong internal performance that is not realized on data from new sites. Even in the absence of recognized confounders, we would caution following Recht et al. that “current accuracy numbers are brittle and susceptible to even minute natural variations in the data distribution” \cite{Recht2018-xc}.  

\bigskip

\noindent Furthermore, high-resolution radiological images are frequently aggressively downsampled (e.g., to 224 x 224 pixels) to facilitate transfer learning, i.e. fine-tuning CNNs pretrained to ImageNet \cite{Huang2016-ct}. While practically convenient, these low-resolution pretrained  models are not optimal for the radiological context because they may eliminate important details in imaging, and we believe the loss of valuable radiographic findings may lead to an increased reliance on confounding factors in making predictions. CNN architectures designed specifically to accommodate the higher resolution of radiological imaging have demonstrated promising early results \cite{Gale2017-ae,Geras2017-fj}.
Given the significant interest in utilizing deep learning to analyze radiological imaging, our findings should give pause to considerations of rapid deployment without thorough vetting of models. No prior studies have assessed whether radiological CNNs generalized to external datasets, which is particularly concerning as there are numerous, protocolized factors that can significantly skew the features in a given radiological image. 

\bigskip

\noindent Even the development of customized deep learning models that are trained, tuned, and tested with the intent of deploying at a single site are not necessarily a solution that can control for potential confounding variables. At a finer level, we found that CNNs could separate portable radiographs from the inpatient wards and emergency department in MSH data with 100\% accuracy, and that these patient groups had significantly different prevalences of pneumonia. It was determined after the fact that devices from different manufacturers had been used in the inpatient units (Konica Minolta) and emergency department (Fujifilm), and the latter were stored in PACS in an inverted color scheme (i.e., air appears white) along with distinctive text indicating laterality and use of a portable scanner. While these identifying features were prominent to the model, they only became apparent to us after manual image review. If certain scanners within a hospital are used to evaluate patients with different baseline disease prevalences (e.g., ICU versus outpatient), these may confound deep learning models trained on radiological data. Fully external testing -- ideally on a collection of data gathered from a varied collection of hospitals -- can reveal and account for such sampling biases that may limit the generalizability a model. 

\bigskip

\noindent While we have focused our analysis on examining degradation of model performance on external test sets, we note that it is possible for external test set performance to be either better or worse than internal. Many different aspects of dataset construction (e.g., inclusion criteria, labeling procedure) and the underlying clinical data (pathology prevalence and severity, confounding protocolized variables) can affect performance. For example, a model trained on noisily-labeled data that included all available imaging might reasonably be expected to have lower internal test performance than if tested externally on a similar dataset manually selected and labeled by a physician as clear examples of pathological and normal cases.

\bigskip

\noindent In addition to site-specific confounding variables that threaten generalizability, there are other factors related to medical management that may exist everywhere but undermine the clinical applicability of a model.  As has been noted, chest drains that treat pneumothorax frequently appear in studies positive for pneumothorax in NIH data; a CNN for pneumothorax may learn to detect obvious chest drains rather than a more subtle pneumothorax itself and might inaccurately negatively diagnose patients presenting with pneumothoraces because they lacked a chest drain \cite{Oakden-Rayner2017-zq}. Ultimately, if CNN-based systems are to be used for medical diagnosis, they must be tailored to carefully considered clinical questions, prospectively tested at a variety of sites in real-world use scenarios, and carefully assessed to determine how they impact diagnostic accuracy.

\bigskip

\noindent There are several limitations to this study. Most notably, without more granular details on the underlying patient populations, we are unable to fully assess what factors might be contributing to the hospital system-specific biasing of the models. The extremely high incidence of pneumonia in the MSH dataset is also a point of concern; however, we attribute this to differences in the underlying patient populations and variability in classification thresholds for pathology. First, a majority of MSH radiographs were portable inpatient scans, ordered for patients too unstable to travel to the radiology department for a standard radiograph. In contrast, all IU radiographs were outpatient. While the inpatient/outpatient mix from NIH is not reported, we believe it likely contains a substantial outpatient percentage given that the incidence of pneumonia is similar to IU. Second, our NLP approach for MSH assigned positive ground truth labels more liberally than NIH or IU, marking a study as positive for pathology when a radiologist explicitly commented on it as a possibility in a report, indicating that the radiographic appearance was consistent with the finding. Different radiologists may have different thresholds at which they explicitly include a possible diagnosis in their reports.  Researchers working in this area will continually have to make decisions about their classification threshold for labeling a study positive or negative. We believe that either of these two factors can drive large differences in prevalences of pathology across datasets, and this variation can confound diagnostic CNNs.

\bigskip

\noindent An additional limitation was that radiologic diagnoses are made in the context of a patient’s history and clinical presentation, something not incorporated into our approach. Positive findings on chest radiograph are necessary but not sufficient for the diagnosis of pneumonia, which is only made when the patient also exhibits a “constellation of suggestive clinical features” \cite{Mandell2007-zg}. Modeling approaches that combine clinical data with imaging findings, reflecting how radiologists practice, may be able to elucidate the contribution of each piece of information and offer more informative predictions. Finally, the relatively small size and low number of pneumonia cases in Indiana data led to wide confidence intervals in IU test AUC and may have limited our ability to detect external performance degradation in some cases. Nevertheless, many key comparisons achieved statistical significance with even this smaller external dataset.

\section*{Conclusion}

Pneumonia screening CNNs achieved better internal than external performance in 3 / 5 natural comparisons. When models were trained on pooled data from sites with different pneumonia prevalence, they performed better on new pooled data from these sites but not on external data. CNNs robustly identified hospital system and department within a hospital which can have large differences in disease burden and may confound disease predictions.

\bibliography{cxr_generalize}



%
%
%

\end{document}